# Effective Data Mining Technique for Classification Cancers via Mutations in Gene using Neural Network

Ayad Ghany Ismaeel
Information System Engineering Department
Technical Engineering College, Erbil Polytechnic University
Erbil, Iraq

Dina Yousif Mikhail
Information System Engineering Department
Technical Engineering College, Erbil Polytechnic University
Erbil, Iraq

*Abstract*—The prediction plays the important role in detecting efficient protection and therapy/treatment of cancer. The prediction of mutations in gene needs a diagnostic and classification, which is based on the whole database (big dataset enough), to reach sufficient accuracy/correct results. Since the tumor suppressor P53 is approximately about fifty percentage of all human tumors because mutations that occur in the TP53 gene into the cells. So, this paper is applied on tumor p53, where the problem is there are several primitive databases (e.g. excel genome and protein database) contain datasets of TP53 gene with its tumor protein p53, these databases are rich datasets that cover all mutations and cause diseases (cancers). But these Data Bases cannot reach to predict and diagnosis cancers, i.e. the big datasets have not efficient Data Mining method, which can predict, diagnosis the mutation, and classify the cancer of patient. The goal of this paper to reach a Data Mining technique, that employs neural network, which bases on the big datasets. Also, offers friendly predictions, flexible, and effective classified cancers, in order to overcome the previous techniques drawbacks. This proposed technique is done by using two approaches, first, bioinformatics techniques by using BLAST, CLUSTALW, etc, in order to know if there are malignant mutations or not. The second, data mining by using neural network; it is selected (12) out of (53) TP53 gene database fields. To clarify, one of these 12 fields (gene location field) did not exists inTP53 gene database; therefore, it is added to the database of TP53 gene in training and testing back propagation algorithm, in order to classify specifically the types of cancers. Feed Forward Back Propagation supports this Data Mining method with data training rate (1) and Mean Square Error (MSE) (0.00000000000001). This effective technique allows in a quick, accurate and easy way to classify the type of cancer.

*Keywords—Detection; Classification; Data Mining; TP53 Gene; Tumor Protein P53; Back Propagation Network (BPN)*

## I. INTRODUCTION

Cancer is a main cause of death worldwide; it has calculated for 7.4 million deaths in 2004 with an estimated 12 million deaths in 2030 [1]. Tumor protein P53, which is produced by Tumor Protein (TP53) gene, is a sequence-specific transcription factor that acts as a large tumor suppressor in mammals. The disorder in the function of the tumor suppressor p53 is one of the most common genetic changes in human cancer, which is close to 50% of all human tumors carry p53 gene mutations within their cells [2]. Fig. 1 shows the cancers and TP53 mutations on the worldwide.

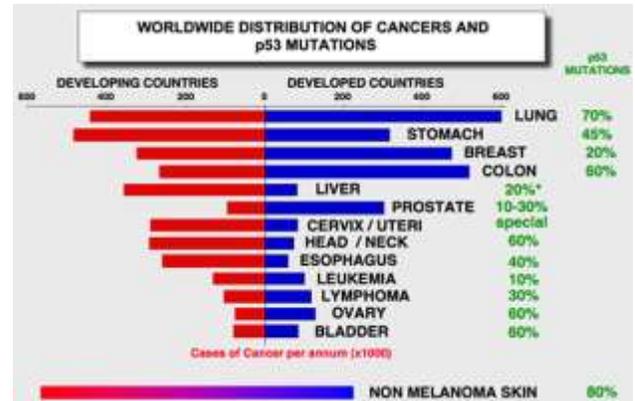

Fig. 1. Shows P53 (TP53 gene) mutations

Nowadays, biologists use a computer system like any other trained professionals but in general function. In addition, they use computers to solve problems that are very specific to them in the specialist tasks. They are taken together, to support the field of bioinformatics. More specifically, bioinformatics' focus is to analyze biological data and to do anticipations about biological systems, in order to provide more knowledge about how living organisms function [3]. Bioinformatics is an emerging discipline that bases upon the strengths of computer sciences, mathematics, and information technology to determine and analyze genetic information [4].

For instance, to predict whether two proteins react or not, it could be used computational biology. If the prediction is correct, then biological data that got from a wet lab experiment, including the proteins, should be analyzed by using computational biology to know how these proteins contribute to the physiology of an organism. Computational biology can be further broken down into molecular modeling and bioinformatics [3].

Data Mining (DM) is defined as the essence of the Knowledge Discovery in Databases (KDD) process. It includes the algorithm conclusions that explore the data, develop the model, and discover previously unknown models. The model is applied to understand phenomena from the data analysis and prognosis. The accessibility and abundance of data today makes knowledge discovery and Data Mining an issue of great necessity and importance [5].

At last, Data Base (DB) related to tumor protein P53 (TP53 gene) contains large amounts of data, these data in the database are represented as excel sheet file, and regular techniques may not be helpful and impractical in such large volumes of data. So, artificial intelligence techniques such as Data Mining are used to simplify and improve the process of research and education. Data Mining is the method of analyzing the data by linking them with artificial intelligence techniques to examine and search for specific information, in addition, to take the useful data from a large amount of data. Moreover, Data Mining can be done through the process of linking this data analysis and artificial intelligence methods to become the most. This proposed method predicts, diagnose, and Classifies cancer mutations; So, the comparison between a person and standard gene protein sequence is done firstly by using BioEdit package. If differences might be found between the two proteins, then there is a malignant mutation. This stage has been used by bioinformatics techniques. Secondly, Data Mining technique (back propagation algorithm) is trained by using UMD Cell-line-2010 p53 mutation database that must be carefully selected to function correctly.

Artificial Neural Network (ANN) can discover how to solve problems by itself. Later, this trained Back Propagation Neural Network (BPNN) offers an effective and flexible predictions, diagnosis and genetic diagnosis technique of cancers.

## II. RELATED WORK

E. Adetiba, J. C. Ekeh, V. O. Matthews, and et al. [2011], this proposed study is aimed at assessing the best back-propagation learning algorithm for a genomic-based on ANN system for NSCLC diagnosis. It used the nucleotide sequences of EGFR's exon 19 of a noncancerous cell to learn ANN. MATLAB R2008a was used to test many BPNN training algorithms to get an optimal algorithm for learning the network. It were examined in the nine different algorithms and achieved the better performance (i.e. the least Mean Square Error MSE) with the minimum epoch (training iterations) and learning time using the Levenberg-Marquardt algorithm (trainlm) [6].

Syed Umar Amin1, Kavita Agarwal, and et al. [2013], introduced a new method to predict heart disease based on the neural network and genetic algorithm. The whole existent systems predicted heart diseases that depended on the clinical dataset, which is collected from complex tests that conducted in pathology labs. There was no method, which predicts heart diseases that depended on risk factors like diabetes, age, family history, high cholesterol, alcohol intake, tobacco smoking, obesity or physical inactivity etc. However, this system gave a patient a warning about a probable existence of heart disease even before he/she makes medical checkups. Two Data Mining tools, genetic algorithms and neural networks were used in this system. In this method, the system may not fall into the local minimum, because the genetic algorithm was applied for optimization of neural networks weights. This system used a multilayered feed-forward network with structure 12 nodes in the input layer, 10 nodes in the hidden layer and 2 nodes in the output layer, where the number of input nodes depends on the final set of risk factors for each patient. In the initial stage, the 'configure' function that available in MATLAB was used to initialize the neural network weights. After that, "these configured weights were passed to the genetic algorithm for optimization according to the fitness function". Once the weights were optimized, the 'trainlm' back propagation algorithm was used for training and learning. The accuracy of the system that predicts heart disease risk is 89%, because the learning process of the derived system was quick, more steady, and accurate as compared to back propagation neural network [7].

Ayad. Ghany Ismaeel, and Raghad. Zuhair Yousif [2015], proposed technique to classify, diagnose mutations' patient, and predict the mutation's position for the patient. TP53 gene (tumor protein P53) datasets were used and (6) fields were selected from UMD_Cell_line_2010 database, in order to train and test Quick back Propagation Network (QPN).The mining method was based on training (QPN), which is an improvement of the back propagation network, since (283-141-1) the number of nodes were used in input, hidden and output layers, by Alyuda NeuroIntelligence package. The training for all datasets (train, test, and validation dataset) led to the following results: the Correlation (0.9993), R-squared (0.9987), and mean of Absolute Relative Error (0.0057) [8].

## III. PROPOSED OF EFFECTIVE DATA MINING METHOD FOR CLASSIFICATION CANCERS

The major functions of suggested Effective Data Mining technique for classifying specific cancer are shown in Fig. 2. The technique of classification specific cancer is done by using two approaches. The first approach predicts whethere the person has mutations that cause cancer or not. The second approach tha classifies the mutations are obtained from the first approach to know which kind of cancer it caused(cancers types) . Those two approaches are:

*A. Bioinformatics Techniques:*

There are many bioinformatics techniques for analysis and search genome, some of these helpful techniques are explained below:

*1) BLAST:* "Is defined as A powerful tool for searching sequence databases with an implementing sequence. BLAST is Basic Linear Alignment Sequence Tool. An earlier program, BLAST, worked by identifying local regions of similarity without gaps and then combining them together. BLAST includes an iterative process, as the emergent pattern becomes better defined in sequential stages of the search"[9].

*2) CLUSTALW*: It is the first technique for examining whether the person has a malicious mutation or not, which is based on the idea of "Two proteins can have very different amino acid sequences, it still be biologically similar (Homology)" [10]. The gene mutation is detected by using CLUSTALW. The gene mutation increases the probability of cancer. In CLUSTALW, users must know there are two types of sequence: one of them is the normal sequence of each gene (without mutation), the second is the person's gene sequence. The matching between them is examined [11].

The previous studies [11, 12] supported the following algorithm, which clarifies the major functions of the bioinformatics tools (sequences alignment).

**Input:** Standard Gene and persons TP53 Gene Sequence
**Output**: Diagnose are there malicious mutations or not in Person's P53 sequence.
**BEGIN**
    Step1: Make FASTA format of Standard Gene and persons Gene Sequence
    Step2: Use ClustalW for Sequences Similarity Check
    Step3: If there is matching
        Its normal gene
    Step4: Else
        Convert Gene Sequence from DNA to Protein
    Step5: Apply ClustalW for Protein Similarity Check
    Step6: If there is matching
        Its normal gene
    Step7: Else
        There is Malignant Mutations.
**END**

enough to classify a specific type of cancer. So, this proposed method needs the second approach, which focuses on training back-propagation neural network, because mutations which are related to cancer and mutations are gotten from the first approach need to be classified by neural networks. Learning or training stage is done by applied The Levenberg-Marquardt back propagation algorithm 'trainlm' is a network learning function that updates weight and bias values according to Levenberg-Marquardt optimization. The proposed structure of training BPN has 3 layers (input layer within layers of BPNN).

"The BPNN is a multilayered neural network applies a supervised learning method and feed-forward architecture. It is by far the most extensively used network". Classifying and predicting are done by using BPNN, because it is one of the most frequently used neural network techniques. "The principle of BPNN runs by approximating the non-linear relationship between the input and the output by adjusting the weight values internally". The neural network model is constructed by using the supervised learning algorithm of back propagation. [13].

The feed forward BPNN is a very common model in neural networks. The errors are back propagated during training, because it does not have feedback connections [14]. The Back-propagation learning process includes two stages in all different layers of the network: forward pass and backward pass [15]:

Forward pass: Input vector is entered to the sensory nodes of the network and its effect spreads out through the network, layer by layer. Lastly, a set of outputs is generated as the actual reaction of the network. During the forward pass, the synaptic weights of the network are all steady.

Backward pass: "The synaptic weights are all modified in accordance with an error correction base". An error signal can be computed by subtracted the actual response of the network from the desired response. "Then the error signal is back propagated through the network, against the direction of synaptic connections".

The steps of training BPN are shown below [16]: The terminologies needed in the algorithm are explained below:

$x_i$ – Input value

$v_{ij}$ – input weight of hidden node

$v_{0j}$ – Weight of bias node from input to hidden

$z\_in_j$ – Weight from input to hidden node

$Z_j$ – output weight of hidden node

$W_{jk}$ – bias node weight from hidden to output

$W_{ok}$ – bias node weight from hidden to output

$y\_in_j$ – input to hidden node

$y_j$ – Final output value

During the forward pass, information passes from the input node to the hidden node, until reaching to the output node. "All input nodes in the input layer are loaded with the values that are given for training. And for each input pattern, a target

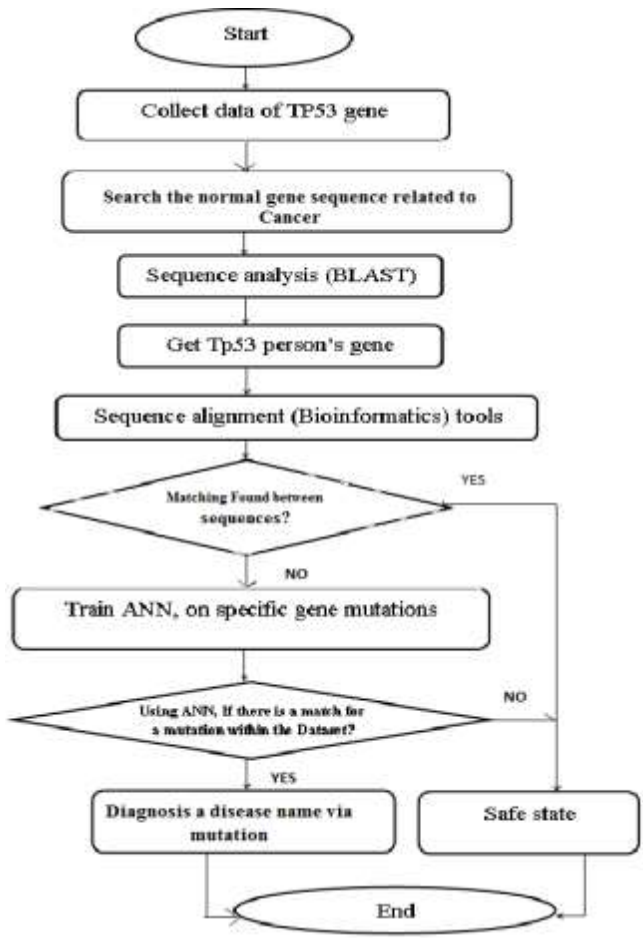

Fig. 2. Flowchart for main tasks of proposed Data Mining method

*B. Data Mining technique (Feed Forward Back Propagation Neural Network):*

After the first approach predicts there is a malignant mutation or not in the person's genes, then these results which obtained from bioinformatics technique(first approach) are not output is also supplied. Each hidden node sums up all incoming values and its bias and then is passed to an activation function f(x)".

$$Z_{\_in_j} = v_{oj} + \sum_{i=1}^{n} x_i \ v_{ij} \quad (1)$$

$$z_j = f(z_{\_in_j}) \quad (2)$$

The output value is passed from the hidden node to each output node. The value from each hidden node is summed up by the output node, and then the output value passes to activation function.

$$y_{\_in_k} = w_{ok} + \sum_{i=1}^{p} z_j \ w_{jk} \quad (3)$$

$$y_k = f(y_{\_in_k}) \quad (4)$$

Determining the error is the start of the backward pass phase. The difference between the target and actual value represents the error. This error is back propagated to each hidden node. "In order to find that it is passed through the derivative of activation function".

$$\delta_k = (t_k - y_k) f'(y_{\_in_k}) \quad (5)$$

Once is found as $\delta_k$, the change in weight can be easily computed

$$\Delta w_{jk} = \alpha \delta_k z_j \quad (6)$$

$$\Delta w_{ok} = \alpha \delta_k \quad (7)$$

Learning rate determines how fast the model learns. If the Learning rate sets to a small value, then the network will need a long time to learn, but if it sets to a high value, then it will make the network inefficient "when there are variations in the input pattern. Updating the weights between the input and hidden layers require more calculations".

$$\delta_{\_in_j} = \sum_{k=1}^{m} \delta_k \ w_{jk} \quad (8)$$

$$\delta_j = \delta_{\_in_j} f'(z_{\_in_j}) \quad (9)$$

$$\Delta v_{ij} = \alpha \delta_j \ x_i \quad (10)$$

$$\Delta v_{oj} = \alpha \delta_j \quad (11)$$

To obtain the updated weights, the old weights are added with the change.

$$w_{jk}(new) = w_{jk}(old) + \Delta w_{jk} \quad (12)$$

$$w_{ok}(new) = w_{ok}(old) + \Delta w_{ok} \quad (13)$$

The process is repeated until the selected error criterion is satisfied.

## IV. EXPERIMENTAL RESULTS

The Implementation of the Effective Data Mining Technique for classifying cancers via mutations in the gene (Tp53) is explained below:

*1)* First, the normal TP53 gene sequence is obtained from The Catalogue of Somatic Mutations In Cancer (COSMIC) site. It provides normal genes, genes information, and datasets. The search for the normal gene can be done by sending gene's name to the server, then selecting the sequence option which provides access to the normal gene (Deoxyribonucleic acid (DNA) or protein sequence).

*2)* It uses BioEdit package to get TP53 gene for person by selecting World Wide Web, then selecting BLAST at the National Center for Biotechnology Information (NCBI), after that selecting nucleotide blast. Then the gene sequence is pasted or uploaded from the file of the normal TP53 gene sequence, which is formatted in FASTA file.

*3)* The Effective Data Mining method uses BioEdit package to complete the first approach for prediction and diagnosis mutations. It applies clustalW to display alignment result between the normal gene and person's gene sequences. A comparison between normal's gene sequences (e.g. Tp53) with the person's gene is executed to find whether there are mutations in person's gene or not, as shown in Fig.3.

*4)* Step (3) is not enough, because its result cannot determine whether mutation affects in protein function or not. So, normal and person TP53 gene are transformed to tumor protein P53. Then, the same tool ClustalW in BioEdit package is used in order to diagnose whether there is malignant mutation as it is done in step (3) or not (No risk). Fig. 4 shows there is malignant mutation (ACC → CCC), i.e. the alignment finds the codon 155 converted from T (at Normal P53) to P (at Person's P53 gene).

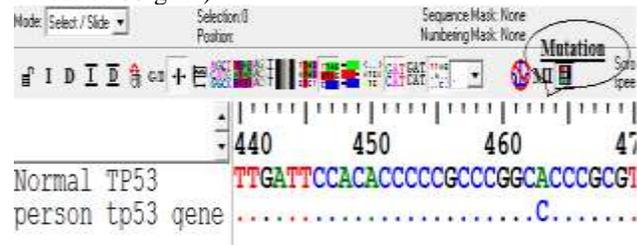

Fig. 3. Shows there is a malignant mutation in TP53 gene

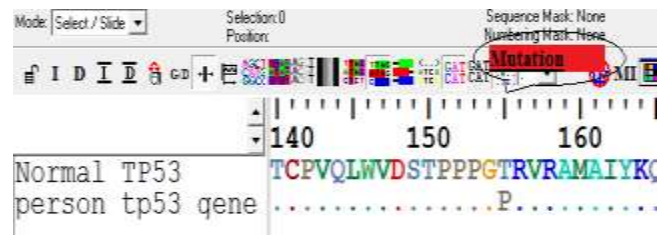

Fig. 4. Shows there is a malignant mutation in P53 protien

*5)* The previous step is used only to detect and predict malignant mutations. Also it does not give the cancer classification results, because the malicious mutations, which are discovered, are general. These malicious mutations related to Tp53 gene database. The common database (UMD_Cell_line_2010) is used to train BPN, which consists of (53) fields and 1448 records. The database (UMD_Cell_line_2010) from TP53 website which is modern and comprehensive database under URL:http://p53.free.fr/Database/p53_MUT_MA T. html[2]. But in Effective Data Mining method, (12) fields are selected for training and testing BPNN; (11) fields are selected from UMD_Cell_line_2010 database. The remaining field is a new field called (gene location field), which is added to the (11) fields selected, in order to get accurate and efficient results in cancer classification. The sample of this database is shown in Fig. 5.

*6)* Matlab R2015a is used on PC type core i5 for neural network toolbox, because it contains several tasks. The classification of malicious mutations for cancer is created successfully by using the structure of feed-forward BPNN and (trainlm) algorithm to obtain an optimal classifier for classification cancer with MSE (0.1E10-13) as shown in Fig. 6.

Fig. 5. Shows sample of dataset (database) which used in training BPN

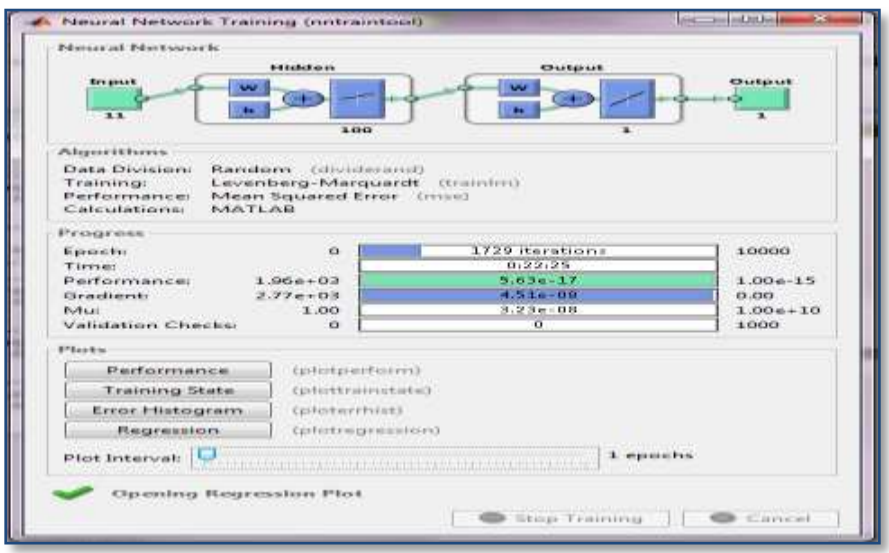

Fig. 6. Shows training result of BP algorithm

Fig. 7 shows plots and the elements of this learning process (Fig. 7; A reveals performance, Fig. 7; B shows regression and Fig 7; C reveals training state).

*7)* The trainer BPNN with malignant mutations of TP53 is completed. Then designed GUI for the doctors, biologists and other users of proposed method is tested. The Effective Data Mining method allows to classify cancers via mutations of a certain person (by entering each field of data manually in GUI).

*8)* The malignant mutation at codon 155 (ACC CCC) is obtained from using ClustalW in BioEdit package. Then trainer BPNN can be used to classify cancer type that may occur due to this mutation, for example, the result of classifying the malignant mutation at codon 155 (ACC CCC) is Head and Neck SCC Cancer, as shown in Fig.8.

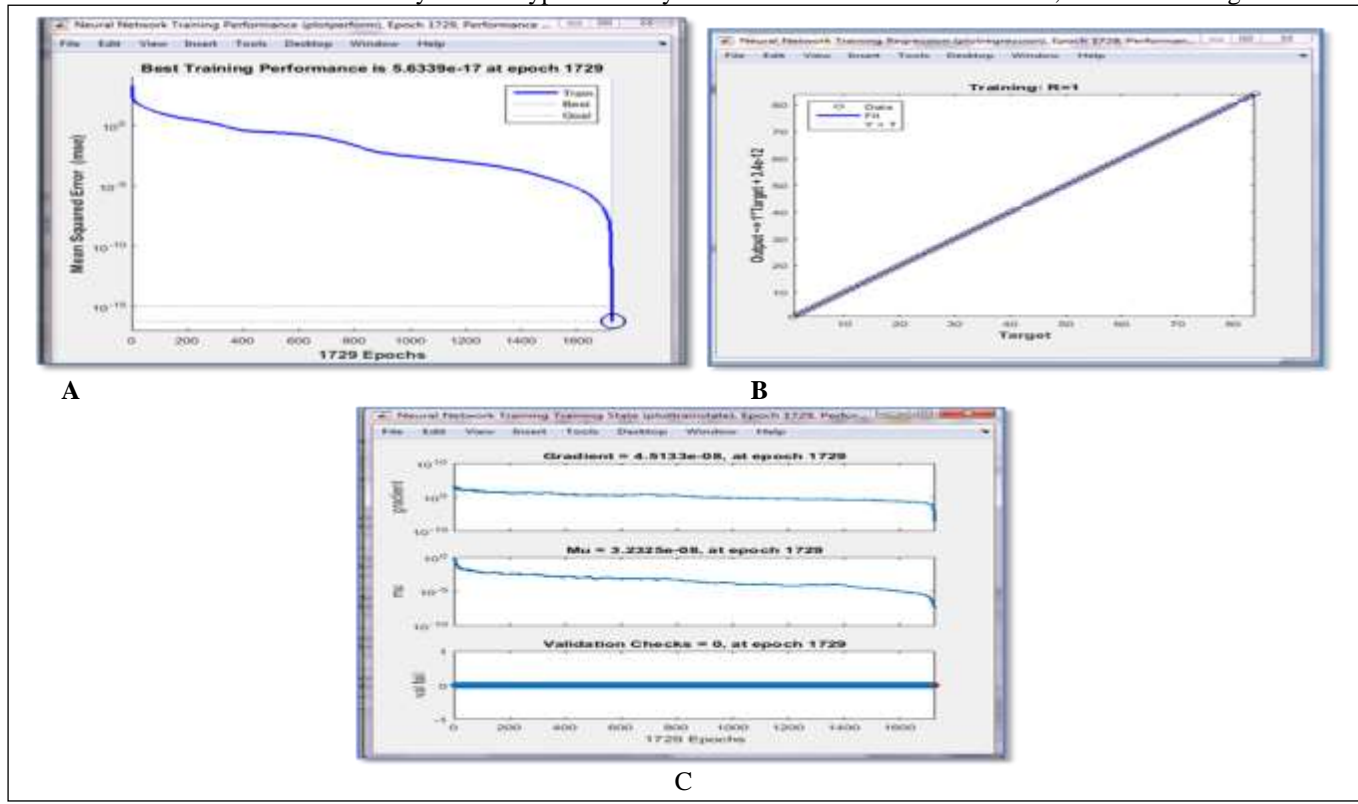

Fig. 7. Shows plots and the three training BPNN elements

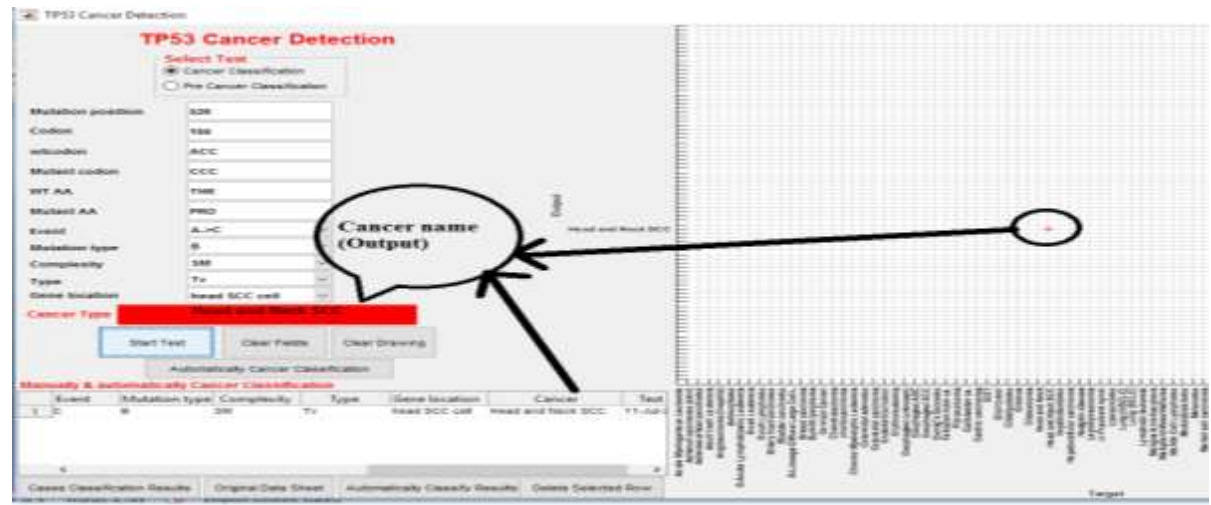

Fig. 8. Classfy cancers via mutations of Person's P53

## V. DISCUSSION THE RESULTS

The learning process is achieved, and it is highly successful. To meet the performance goal, it only takes 22 minutes to complete the learning process. Then the problem is presented to the trained model to classify the cancers. The DM method is an effective way in the classification cancers via mutation, since BPN is used in training and testing a minimum number of fields, which is (12) out of (53) fields in each record of TP53 database. The data of p53 database was saved in columns and records in Excel sheet file, as shown in Fig. 5. Whereas, (7) out of (53) fields for each record of TP53 database were used in the Novel Mining method, and 14 fields were used in the heart diseases method. These fields depend on the final set of risk factors for each patient. In addition, the proposed DM method adds a new field called Gene Location field to the UMD TP53 database in order to make the neural network be able to classify specific type of cancer, and give accurate results; While the other related methods base only on the original database. Moreover, the proposed DM method of classifying cancer is compared with other related techniques or methods. The comparison is done in terms of goals, the used database, neural network structure, techniques, and performance; While all the other related methods use neural network but with different structure, training algorithm, performance, and results. More details are shown in Table I.

TABLE I. SHOWS COMPARISON OF EFFECTIVE DATA MINING METHOD WITH TWO OTHER METHODS

| Feature | The proposed method | Amin, et. al., [7] | Ayad, et. al., [8] |
|---|---|---|---|
| The goal | Prediction, diagnosis ,classification specific cancer | prediction ,diagnosis heart diseases | Prediction, diagnosis ,classification mutation position |
| Universal method | Yes with two approaches, and adding new data field to DB | no | yes |
| Including Tp53 gene | yes | no | yes |
| DNA and Protein check | yes | no | yes |
| Sequence similarity check | yes | no | Yes |
| Data base used | UMD TP53 mutation DB | American Heart Association survey | UMD TP53 mutation DB |
| Technique used | Bioinformatics and DM tools(BPN algorithm) | DM techneques (neural networks and genetic algorithms) | Bioinformatics tools, quick BPN algorithm |
| Weight update function | trainlm | trainlm | QBP |
| ANN topology | (11-100-1) | (12-10-2) | (283-141-1) |
| Performance | 0.1E10-13 | 0.034683 | 0.000006 |
| Program | MATLAB R2015a | MATLAB R2012a | Alyuda NeuroIntellige-nce |
| Support for | Researchers, bioinformatics doctors, and biomedical Eng. | Doctors | Researchers |

## VI. CONCLUSIONS

The proposed Data Mining method of cancer classification explains the following conclusions:

*1)* The proposed Data Mining method provides flexible diagnosis and prediction. Also, it classifies cancers via mutations in tumor protein P53 sequence. BBNN algorithm is used with the best performance (MSE), which reaches to (5.6339E-17), and the training rate(R) equals (1), as shown in Fig.7A&C. While in the Novel Mining method, Quick BPN algorithm was used with performance (0.000006), the training rate(R) equals (0.9987). Whereas in the heart diseases method, BP neural network and genetic algorithms were used with performance (0.034683).

*2)* The proposed approach shows the classification of cancer via predicting mutated P53 gene, in order to reduce the risk of cancer infection. This is done to keep people away from radiation, exposure to toxins, control themselves at older ages, and arrange their food system. In addition, the earlier diagnosis can predict the therapy for the mutated tumor protein P53.

*3)* Since cancer is an inherited disease and many different cases have been appeared in many families history around the world, this study is important for a further work to set up a database for a local area (e.g. for Middle East). This database would include the background or history of each family datasets, by taking into consideration the genetic diseases data of the family history. Then a complete system would constructed that is able to predict genetic disease early. Also, this local database would support therapy process by using therapy techniques like replacement, drug discovery, etc.

*4)* The results are obtained from this method can be forwarded to include the gene therapy by using therapy techniques as it is mentioned in the previous point, where therapy is a field in Biotechnology science.

ACKNOWLEDGMENT

(Mr. HAIDER HASAN HUSSIEN) Thanks for helping me when adding a new field that called gene location field in UMD database and yours advice at the medical side.

## AUTHOR PROFILE

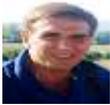

Ayad Ghany Ismaeel is a professor at 7 Aug, 2012 and awarded his Ph.D. computer science in the qualification of the computer from University of Technology, Baghdad- Iraq at 2006. M.Sc. in computer science (applied) from the National Computers Center NCC (currently ICCI) Baghdad-Iraq at 1987, and then B.Sc. in Informatics/ statistics from Al-Mustansiriyah University, Baghdad- Iraq at 1982.

Professor Ismaeel is coordinator and organizer of computer/IT center in Baquba Technical Institute - Technical Education Foundation TEF Baghdad-Iraq at 1990, founder and coordinator the dept. of Computer Systems. in Baquba Technical Institute- TEF Baghdad-Iraq at 2000, then founder and coordinator the department of Information Systems Engineering, Erbil Technical Engineering College - Erbil Polytechnic University, Iraq at 2007. He has 28-year experience in teaching an undergraduate and graduate in computer science, information systems, software engineering and fields related (IT, bioinformatics, biomedical Engineering/Informatics, etc) in many universities from 2007 till now in Kurdistan, Iraq. He is editorial, advisor, reviewer board member (one of them IJACSA of SAI Org:: http://thesai.org/Reviewers/Details/0a1f2c5d-6c63-4232-9fa2-d790812be480 ) and program committee member of many international journals and conferences. His research interest mobile network, cloud computing, semantic web, distributed system, healthcare systems bioinformatics & biomedical Eng. /Informatics. He has experiences and skills for Advising, Counseling, Teaching, Training, Industrial and Curricula Development using European/Germany standards. More details visit: http://drayadghanyismaeel.wix.com/ayad-ghany-ismaeel-

**Dina Yousif Mikhail**: obtained B.Sc. (Bachelor of engineering) Medical Instrumentation engineering at 2003/2004 in Technical Collage - University of Mosul-Iraq and Higher Diploma (Software Engineering) at 2009/2010 in Engineer Collage-University of Sallah Al Dien, Erbil-Iraq. She is M.Sc. researcher's student in Information System Engineering in Engineering Technical Collage - Erbil, Polytechnic University Erbil-Iraq. She is currently in the department of Information Systems Engineering, Technical Engineering College, Erbil Polytechnic University, IRAQ. Her research interests in web application, bioinformatics & biomedical Engineering, Artificial intelligent. She has 10-year experience in teaching an undergraduate in information systems engineering.